\pdfoutput=1
\documentclass[letterpaper, 10 pt, conference]{ieeeconf}  

\IEEEoverridecommandlockouts                              

\usepackage{booktabs}
\usepackage{array}
\usepackage{graphicx}
\usepackage{xcolor}
\usepackage{tikz}
\usepackage{cite}
\usepackage{multirow}
\usepackage{amssymb}
\usepackage{eucal}
\usepackage{caption}
\usepackage{amsmath,amsfonts}
\usepackage{array}
\usepackage{textcomp}
\usepackage{stfloats}
\usepackage{url}
\usepackage{verbatim}
\usepackage{graphicx}
\usepackage{stfloats}
\usepackage{color}
\usepackage{cite}
\usepackage{multirow}
\usepackage{makecell}
\usepackage[T1]{fontenc}
\usepackage{authblk}
\usepackage{rotating}

\makeatletter
\let\NAT@parse\undefined
\makeatother
\usepackage[hidelinks]{hyperref}

\definecolor{Building}{RGB}{255,200,0}
\definecolor{Sky}{RGB}{50,255,255}
\definecolor{Tree}{RGB}{0,175,0}
\definecolor{Road}{RGB}{217,217,217}
\definecolor{Sidewalk}{RGB}{75,0,75}
\definecolor{Person}{RGB}{255,30,30}
\definecolor{Plant}{RGB}{135,60,0}
\definecolor{Car}{RGB}{100,150,245}
\definecolor{Fence}{RGB}{245,230,100}
\definecolor{Signboard}{RGB}{255,50,255}
\definecolor{Bus}{RGB}{30,30,255}
\definecolor{Truck}{RGB}{180,30,80}
\definecolor{Streetlight}{RGB}{255,240,150}
\definecolor{Pole}{RGB}{0,0,255}
\definecolor{Van}{RGB}{250,80,100}
\definecolor{Bicycle}{RGB}{100,230,245}
\definecolor{Trafficlight}{RGB}{90,30,150}
\definecolor{Motorcyclist}{RGB}{170,255,150}

\title{
\LARGE \bf
PointSSC: A Cooperative Vehicle-Infrastructure Point Cloud Benchmark for Semantic Scene Completion
}

\author{Yuxiang Yan$^{1}$, Boda Liu$^{1}$, Jianfei Ai$^{2}$, Qinbu Li$^{2}$, Ru Wan$^{2}$ and Jian Pu$^{*,1}$
\thanks{*Corresponding author, {\tt\small jianpu@fudan.edu.cn}}
\thanks{$^{1}$Institute of Science and Technology for Brain-Inspired Intelligence, Fudan University.
        {\tt\small \{yxyan22, bdliu22\}@m.fudan.edu.cn }}%
\thanks{$^{2}$Mogo Auto Intelligence and Telematics Information Technology Company Ltd.
        {\tt\small \{aijianfei, liqinbu, wanru\}@zhidaoauto.com}
        }%
}

\begin{document}

\maketitle
\thispagestyle{empty}
\pagestyle{empty}

\begin{abstract}
Semantic Scene Completion (SSC) aims to jointly generate space occupancies and semantic labels for complex 3D scenes. Most existing SSC models focus on volumetric representations, which are memory-inefficient for large outdoor spaces. Point clouds provide a lightweight alternative but existing benchmarks lack outdoor point cloud scenes with semantic labels. To address this, we introduce PointSSC, the first cooperative vehicle-infrastructure point cloud benchmark for semantic scene completion. These scenes exhibit long-range perception and minimal occlusion. We develop an automated annotation pipeline leveraging Semantic Segment Anything to efficiently assign semantics. To benchmark progress, we propose a LiDAR-based model with a Spatial-Aware Transformer for global and local feature extraction and a Completion and Segmentation Cooperative Module for joint completion and segmentation. PointSSC provides a challenging testbed to drive advances in semantic point cloud completion for real-world navigation. The code and datasets are available at \textcolor{red}{\href{https://github.com/yyxssm/PointSSC}{https://github.com/yyxssm/PointSSC}}.


\end{abstract}

\section{INTRODUCTION}



Accurate perception of 3D scenes is crucial for autonomous agents to navigate complex environments. Holistic understanding of 3D scenes informs critical downstream tasks such as path planning and collision avoidance. Leading 3D scene perception tasks include 3D object detection, 3D point cloud semantic segmentation, and semantic scene completion (SSC). Similar to how humans can understand 3D environments even when partially hidden, SSC predicts complete geometric shapes and their semantic labels from incomplete data. However, a considerable gap persists between current SSC models and human-level perception for real-world driving scenarios.


Most current SSC datasets rely on vehicle-mounted sensors, which have limited perception range and greater susceptibility to occlusion compared to elevated infrastructure vantage points. SemanticKITTI \cite{c1} provides only front-view semantic scenes, while SurroundOcc \cite{c3} and OpenOccupancy \cite{c4} incorporate surrounding views yet they still do not effectively address occluded areas. Occ3D \cite{c5} uses ray casting to generate occlusion masks, but solely leverages these to refine evaluation metrics rather than improve ground truth labels. Succinctly, existing vehicle-view datasets fail to capture the long-range perception and prevalent occlusion characteristic of real-world driving environments. Purposing a infrastructure-view datasets could enable richer, more complete semantic annotations to further advance SSC research.



\begin{figure}[t]
  \centering
  \includegraphics[width=0.49\textwidth]{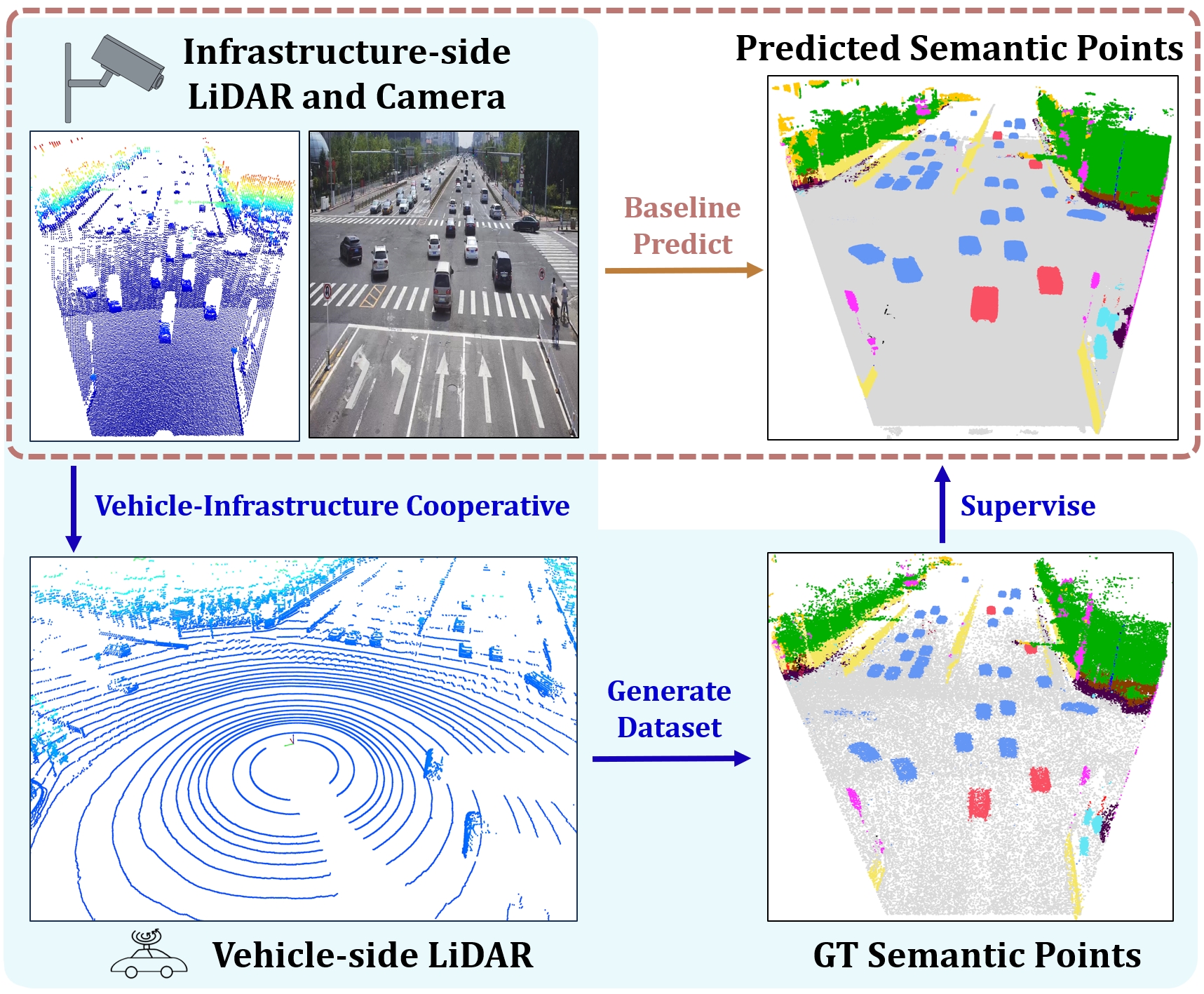}
  \caption{\textbf{PointSSC Overview}. Given infrastructure-side partial points and images (top left), we first couple them with vehicle-side point clouds (bottom left) to construct the PointSSC dataset (bottom right). PointSSC then guides our network (top right) for point cloud semantic scene completion. The \textcolor{blue}{\textbf{blue}} background indicates the PointSSC generation pipeline, while the \textcolor{brown}{\textbf{brown}} dashed box shows model prediction.}
  \label{fig:fig1}
\end{figure}

\begin{table*}[htbp]
  \centering
  \caption{Comparison of PointSSC and Existing Semantic Scene Completion Datasets}
    \begin{tabular}{cccccc p{16.0em}p{11.0em}p{8.0em}p{9.0em}p{9.0em}p{14.0em}}
    \toprule
    Dataset & Type  & Data Volume & Space Range (m) & GT Style & Perspective \\
    \midrule
    SUNCG\cite{c21} & Synthtic & $10^6$ & -     & Voxel & - \\
    NYUv2\cite{c43} & Indoor & $10^6$ & -     & Voxel & - \\
    Occ3D-nuScenes\cite{c5} & Outdoor & $10^5$ & $80\times80\times6$ & Voxel & Vehicle \\
    SurroundOcc\cite{c3} & Outdoor & $10^5$ & $100\times100\times8$ & Voxel & Vehicle \\
    OpenScene\cite{c44} & Outdoor & $10^5$ & $100\times100\times8$ & Voxel & Vehicle \\
    Occ3D-Waymo\cite{c5} & Outdoor & $10^6$ & $80\times80\times1$2 & Voxel & Vehicle \\
    SemanticKITTI\cite{c1} & Outdoor & $10^6$ & $160\times160\times13$ & Voxel & Vehicle \\
    OpenOccupancy\cite{c4} & Outdoor & $10^7$ & $102\times102\times8$ & Voxel & Vehicle \\
    \midrule
    \textbf{PointSSC} (Ours) & Outdoor & $10^7$ & $250\times140\times17$ & Point Cloud  & Vehicle \& Infrastructure \\
    \bottomrule
    \end{tabular}%
  \label{tab:dataset_compare}%
\end{table*}%


To mitigate occlusion affecting vehicle-mounted sensors, we adopt a vehicle-infrastructure cooperative perspective. Infrastructure sensors possess longer range and fewer blindspots, while vehicle sensors enrich scene representation from their distinct vantage. Moreover, compared to volumetric formats, point clouds enable efficient semantic scene representation with minimal memory overhead \cite{c38}. Therefore, we develop PointSSC, the first point cloud semantic scene completion benchmark leveraging cooperative vehicle and infrastructure views, as shown in Figure~\ref{fig:fig1}. PointSSC provides a lightweight yet detailed testbed to advance semantic completion for outdoor autonomous navigation.



In Tab.~\ref{tab:dataset_compare}, we compare our PointSSC dataset against other mainstream semantic scene completion datasets. To the best of authors' knowledge, PointSSC has the largest data volume and spatial coverage. It is the first outdoor point cloud SSC dataset developed cooperatively from vehicle and infrastructure perspectives.
To enable further research, we propose a LiDAR-based model tailored for PointSSC. To handle outdoor point clouds, we introduce a Spatial-Aware Transformer and Completion and Segmentation Cooperative Module (CSCM). Experiments validate the efficacy of these contributions.
Since infrastructure-side perception is becoming more and more important, our model is trained from an infrastructure vantage. Our key innovations are:
\begin{itemize}
\item We present PointSSC, the first large-scale outdoor point cloud SSC dataset from cooperative vehicle-infrastructure views.
\item We propose a baseline model with a Spatial-Aware Transformer and a Completion and Segmentation Cooperative Module.
\item Our method sets the new state-of-the-art on PointSSC for both completion and semantic segmentation tasks.
\end{itemize}

\section{RELATED WORKS}

\subsection{Point Cloud Completion}

Existing point cloud completion methods mostly aim at object-level completion, which can be roughly divided into geometric-based and learning-based approaches. Geometry-based approaches leverage input's inherent geometric structures or template's geometric to infer missing geometric shapes. Such methods\cite{c6,c7,c8,c9,c10,c11} need complex optimization techniques and lack robust generalization capabilities. 

Learning-based approaches employ neural networks for point cloud completion. 
PointNet\cite{c12} and its variants\cite{c13} offered a methodology to directly process unordered point clouds. FoldingNet\cite{c14} and PCN\cite{c16}  pioneered point cloud completion and introduced a two-stage point cloud generation model. SnowFlakeNet\cite{c17} used the growth of neighbor points to complete points. PoinTr\cite{c18} and following works\cite{c19, c20} employed point proxies to reduce computational consumption. A recent work, CasFusionNet\cite{c38} uses a dense feature fusion method to complete semantic point clouds for indoor scenes. Although it is the first scene-level point cloud completion model, it is computation-consuming and cannot perform well in large outdoor scenes. Our PointSSC model explores semantic point cloud completion for large outdoor scenes for the first time and shows exciting results.

\subsection{3D Semantic Scene Completion (SSC)}

Holistic 3D scene understanding plays an important role in autonomous driving perception. However, due to the limitations of sensing devices and viewing angles, it is very challenging.
SSCNet\cite{c21} was the first network proposing to use a single-view depth image as input and constructed an end-to-end model to SSC task. 3DSketch\cite{c23} and AICNet\cite{c24} proposed to use images and corresponding depth to generate semantic scenes. Subsequent works \cite{c39, c40, c41, c42,chen2023generative,yang2022deepinteraction} further designed the indoor scene completion model and achieved better performance. To solve SSC tasks in outdoor autonomous driving scenes, JS3CNet\cite{c25} proposed to use LiDAR point clouds for the first time.
LMSCNet\cite{c26} proposed a lightweight structure combining 2D and 3D convolutions to improve inference speed. SCPNet\cite{c27} applied knowledge distillation to SSC tasks. MonoScene\cite{c28} was the first to accomplish semantic scene completion by using monocular RGB images. VoxFormer\cite{c29} and TPVFormer\cite{c30} have further improved performance on the basis of MonoScene. SurroundOcc\cite{c31} took multi-view camera images as input and used occupancy prediction to predict occupancy semantics. OpenOccupancy\cite{c32} proposed a nuScenes-based semantic occupancy prediction dataset and gave a baseline based on unimodality and multimodality. Although these SSC models achieve surprising results, they require expensive computing and storage resources due to their volumetric representation.

\subsection{Infrastructure-side Datasets}
Autonomous driving datasets play an indispensable role in semantic scene understanding. Infrastructure-side autonomous driving datasets collect point clouds with fixed LiDARs, which are different from vehicle-side LiDARs that moving with vehicles. IPS300+\cite{c35} introduced a large-scale multimodal dataset for infrastructure-side perception tasks in urban intersections.
BAAI-VANJEE\cite{c36} is an infrastructure-side object detection dataset featuring diverse scenes.
DAIR-V2X\cite{c37} and its following work\cite{v2x-seq} are both large-scale vehicle-infrastructure cooperative multimodal datasets.
They can be used for vehicle-infrastructure cooperative perception, prediction and other related tasks. Our dataset is developed based on V2X-Seq\cite{v2x-seq} and is tailored for point cloud SSC tasks.

\section{PointSSC Generation Pipeline}

\begin{figure*}[thpb]
  \centering
  \includegraphics[width=0.96\textwidth]{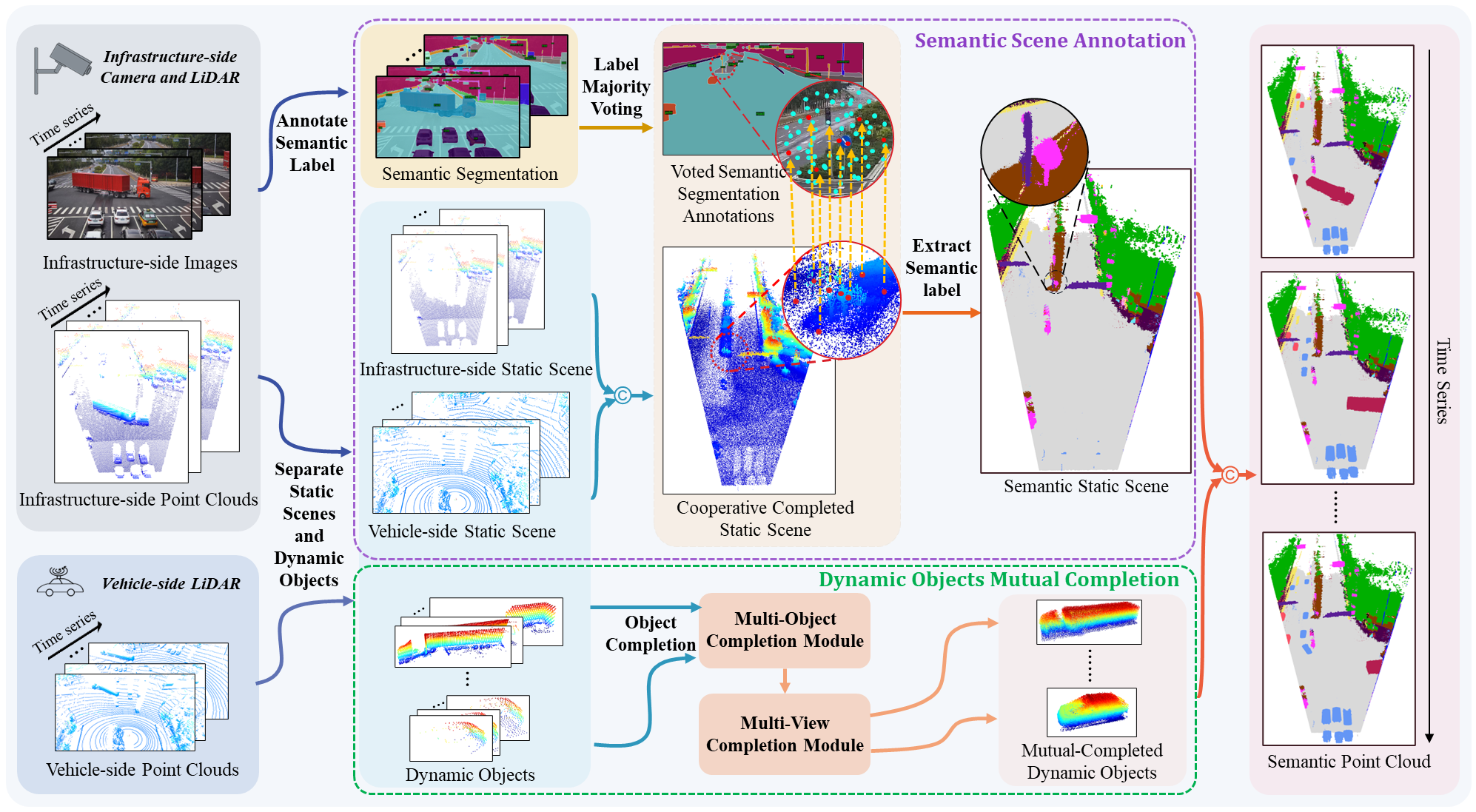}
  \caption{Pipeline of our \textbf{PointSSC} dataset generation. For infrastructure-side images, we annotate their semantic labels. For vehicle-infrastructure cooperative point clouds, we use ground truth bounding boxes to separate static scenes and dynamic objects. For static scenes, we concatenate multi-frame static scenes together and annotate 2D semantic labels to 3D points to get semantic static scenes. For dynamic objects, we use a multi-view, multi-object completion strategy to complete them. Finally, we concatenate semantic static scenes and dynamic objects together. \textcircled{c} denotes the concatenate operation.}
  \label{fig:dataset_pipeline}
\end{figure*}

In this section, we introduce the task definition of point cloud semantic scene completion in \ref{sec:ssc_definition}. Subsequently, the PointSSC generation pipeline will be introduced, including semantic scene annotation in Sec. \ref{sec:semantic_anno} and dynamic objects mutual completion in Sec. \ref{sec:object_completion}.

\subsection{Overview}
\label{sec:ssc_definition}
Point cloud SSC aims to generate complete point clouds and their semantic labels cooperatively. 
Specifically, given  partial point clouds or images, SSC models are required to generate a tuple $\left( {{P},{L}} \right)$, where ${P} \in \mathbb{R}{^{N \times 3}}$ are complete points and ${L} \in \mathbb{R}{^{N \times 1}}$ are corresponding semantic labels. Compared to volumetric SSC, point cloud SSC is more memory-efficiency and has a stronger ability to represent complex scenes.

Our PointSSC benchmark is developed based on sequential vehicle-infrastructure cooperative dataset V2X-Seq\cite{v2x-seq}, which comprises 11,275 frames of both LiDAR and camera frames and annotates dynamic object bounding boxes in 9 classes. To obtain a scene-level point cloud SSC dataset, we propose a pipeline to generate complete scene-level point clouds and semantic labels simultaneously.

Infrastructure-side sensor's sensing range is often several times that of the vehicle sides. However, infrastructure sensors are fixed on the roadside equipment, so they can only perceive the scene from a limited view. As vehicle-side sensors can provide different view information, vehicle-side and infrastructure-side sensors can complement each other. Fig. \ref{fig:dataset_pipeline} shows our PointSSC generation pipeline, we use vehicle-infrastructure collected point clouds and infrastructure-side images to develop ground truth semantic background point (purple dashed box). To handle incomplete perception of dynamic objects, we apply a multi-object multi-view completion strategy to dynamic objects (green dashed box).

\subsection{Semantic Scene Annotation}
\label{sec:semantic_anno}
We divide the semantic scene annotation module into two parts. Firstly, we perform multi-frame static scene completion, then we use 2D image segmentation to annotate 3D static scene point cloud segmentation labels.

\textbf{Multi-frame Static Scene Completion.}
We separate static scenes and dynamic objects according to ground truth bounding box annotations. Then, we use LiDAR extrinsic calibration matrix to transform vehicle-infrastructure cooperative point clouds into a world coordinate system  for registration preparation. 
For static scenes, we directly use concatenate operations to generate a complete static scene.

\textbf{Semantic Point Cloud Annotation.}
Compared to most mainstream SSC datasets\cite{c1,c3,c4,c5,c44} can generate semantic voxels by using existing point cloud semantic segmentation labels, the object detection dataset V2X-Seq\cite{v2x-seq} does not have a semantic annotation of point clouds or images. As it is labor-consuming to annotate semantic segmentation manually, we design an automatic point cloud semantic annotation pipeline based on image segmentation. Note that dynamic objects in V2X-Seq\cite{v2x-seq} are semantically labeled by bounding boxes, we only annotate background points.

Since image semantic segmentation generally shows superior performance than point cloud segmentation and Segment Anything\cite{kirillov2023segany} can adapt to domain gap well, we use Semantic Segment Anything\cite{chen2023semantic} to generate image semantic segmentation labels. To address the problem that background scenes are often occluded by dynamic objects, we propose a semantic label majority voting method to ease the background segmentation noise. 
Specifically, for each pixel, we annotate the definitive background semantic label by using semantic class observed most across time series images.
Finally, we project point clouds onto images through intrinsic and extrinsic calibration transformation, point clouds can fetch their semantic labels by indexing on image semantic segmentation.

\subsection{Dynamic Objects Mutual Completion}
\label{sec:object_completion}


Point clouds are prone to occlusion, including both self-occlusion and external occlusion, and they tend to be sparse at long range. This means that if we only carry out single-view or single-object registration, the objects produced often lack complete shape characteristics.

Complete geometric representation of dynamic objects can be obtained by vehicle-infrastructure cooperative registration. Therefore, we adopt multi-view and multi-object registration methods to complete dynamic objects.




Inspired by BtcDet\cite{xu2022behind}, we develop a heuristic function \(\mathcal{H}\left(A,B\right)\) to assess the disparity between a source object \(A\) and a target object \(B\). A lower score from \(\mathcal{H}\left(A,B\right)\) indicates a higher similarity between \(A\) and \(B\). To begin, we first collect objects that require completion together into a shape bank $\mathcal{B}$. For a given source object \(A\) that necessitates completion, we compute its similarity with every other object in the shape bank \(\mathcal{B}\). The object in $\mathcal{B}$ that exhibits the highest similarity to \(A\) is 
used to complete the source object \(A\).

In practice, we employ the heuristic function \(\mathcal{H}\) to execute shape completion for both infrastructure-side and vehicle-side objects independently. This process is conceptualized as multi-object completion. Subsequently, to obtain fully completed objects, we treat the infrastructure-side objects as source objects and the vehicle-side objects as target objects. This facilitates a multi-view cooperative completion. After obtaining the completed dynamic objects, we assign them with appropriate semantic labels and reintegrate them into static scenes based on their bounding boxes.

\section{MODELS}

\begin{figure*}[thpb]
  \centering
  \includegraphics[width=1\textwidth]{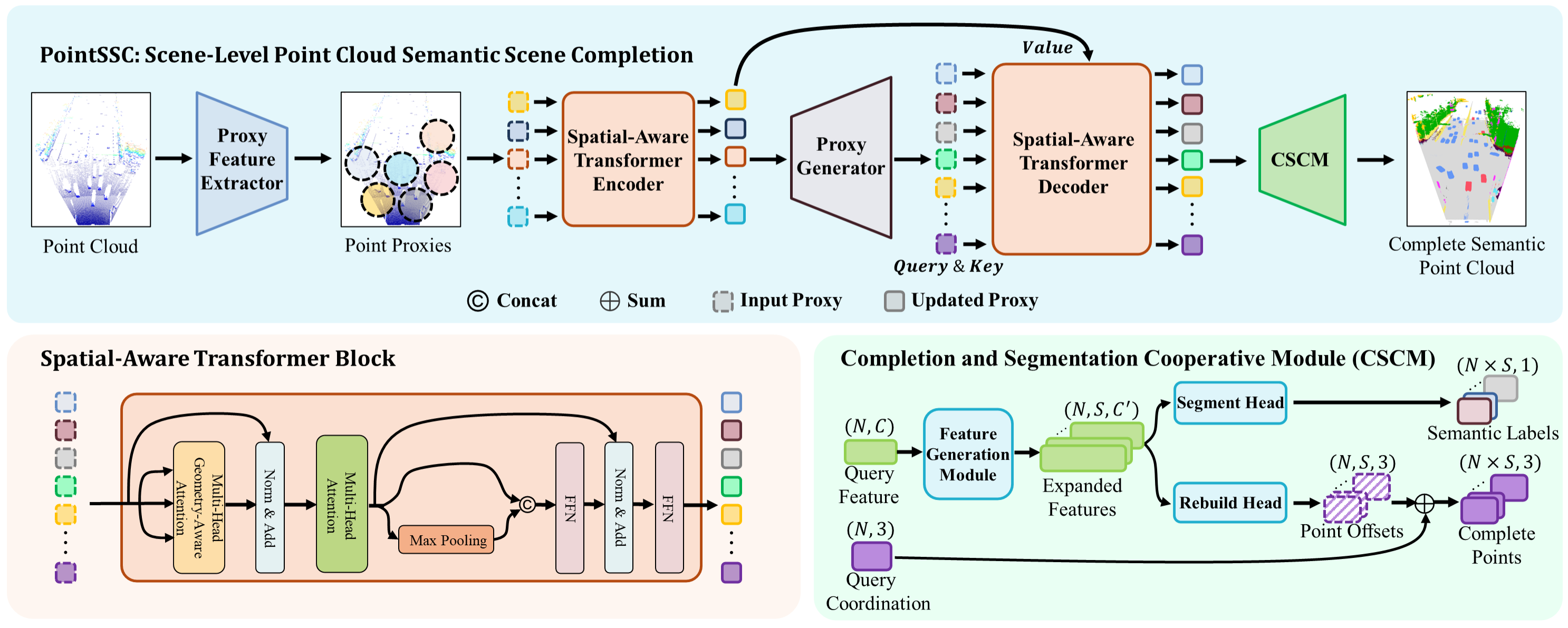}
  \caption{Pipeline of our \textbf{PointSSC} baseline. Given partial point clouds, we use PointNet++\cite{c13} to extract point proxies, then fuse local and global features through a spatial-aware transformer. We use a proxy generator to get coarse-up sampled point proxies and CSCM to generate complete semantic points through a coarse-to-fine strategy.}
  \label{fig:model_pipeline}
\end{figure*}

Fig. \ref{fig:model_pipeline} provides an overview of our model. Firstly, we introduce the overview of our baseline for the PointSSC dataset in Sec. \ref{sec:model_overview}. Secondly, in Sec. \ref{sec:spatial-aware_transformer}, a spatial-aware transformer, which fuses both local and global information effectively will be introduced. Thirdly, in Sec. \ref{sec:model_CSCM} we introduce the completion and segmentation cooperative module (CSCM), which can generate points and corresponding semantic labels cooperatively. Finally, we introduce the training implementation in Sec. \ref{sec:model_training_loss}.

\subsection{Overview}
\label{sec:model_overview}
Point cloud SSC aims to perform scene-level semantic point cloud generation. To facilitate subsequent research and use, we propose a LiDAR-based model as a baseline for PointSSC. 



Our baseline is developed on Transformer\cite{vaswani2017attention} based model AdaPoinTr\cite{c19}. A defining feature of the attention mechanism is that both inputs and outputs are order-independent. This characteristic aligns seamlessly with the inherently unordered nature of point clouds. Moreover, considering the huge computational expense of processing each individual point, instead of upsampling origin point clouds ${P_0} \in {\mathbb{R}^{{N_0} \times 3}}$ directly, we alter to use a combination of points ${P_1} \in {\mathbb{R}^{{N_1} \times 3}}$ along with high-dimensional features ${F_1} \in {\mathbb{R}^{{N_1} \times {C_1}}}$ to encapsulate the entirety of the point cloud, where ${N_1} \ll {N_0}$ and $C_1 \gg 3$. The combination $\{P_1,F_1 \}$ is called as proxy. The synergy of proxies and the transformer architecture enables our networks to adeptly capture local spatial correlations, which is indispensable for the point cloud semantic scene completion task. Guided by these insights, we have fashioned our baseline models.

Given input partial point clouds, we employ a proxy feature extractor to obtain both proxy coordinates and associated features. Then we use a spatial-aware transformer encoder to fuse global and local features, followed by a proxy generator to generate coarse scene points. After that, a spatial-aware transformer decoder is applied to further extract local features. Finally, the Completion and Segmentation Cooperative Module (CSCM) is utilized to collaboratively generate complete points along with their semantic labels.


\subsection{Spatial-Aware Transformer}
\label{sec:spatial-aware_transformer}

Existing object-level point cloud processing models\cite{c19, c12, guo2021pct} often merge per-point features into a single high-dimensional feature via max pooling. However, this approach may not be suitable for outdoor scene-level point clouds, which typically contain a larger number of points and exhibit greater complexity. We argue that a single high-dimensional feature is insufficient to fully represent an entire outdoor scene, as local geometric information also plays a crucial role. To address this, we introduce the spatial-aware transformer, a novel model that can effectively integrate global and local features. 


We propose an integration of both local and global information within transformer blocks. Initially, we utilize the proxy feature $F_{proxy}$ and the learned position embedding $F_{pe}$ to generate queries $Q$, keys $K$, and values $V$. Subsequently, we employ the geometry-aware attention block from PoinTr\cite{c18} to discern the local geometric structure among points. 
Aiming to fuse both global and local features, we employ max pooling to derive the global feature $F_{global}$. This global feature $F_{global}$ is then concatenated with $F_{proxy}$ and subjected to a feed-forward network (FFN), orchestrating a fusion of the local and global features. We also maintain the skip connection in the standard Transformer to execute an element-wise feature fusion of local features, resulting in updated proxy features $Q'$. The spatial-aware transformer block is described as follows:
\begin{equation}
\label{equ:att_input}
Q,K,V = \textbf{W}_{Q,K,V}\left( {{F_{proxy}} + {F_{pe}}} \right),
\end{equation}
\begin{equation}
\mathrm{Attn} = \mathrm{Softmax} \left( {\frac{{Q{K^T}}}{{\sqrt {{d_k}} }}} \right)V,
\end{equation}
\begin{equation}
F_{global} = \mathrm{Maxpool} \left( {\mathrm{Attn}} \right),
\end{equation}
\begin{equation}
Q' = \mathrm{FFN} \left( {\mathrm{FFN}\left( {\left[ {\mathrm{Attn},F_{global}} \right]} \right) + \mathrm{Attn}} \right),
\end{equation}
where $\textbf{W}_Q$, $\textbf{W}_K$ and $\textbf{W}_V$ are learnable parameters, $d_k$ is the dimension of $K$ and $[\cdot,\cdot]$ represents concatenate operation. Using a spatial-aware transformer, we fuse local features twice and global features once,
so that proxy features are dominated by local features while they still retain global scene information,
which is more suitable for our scene-level point cloud generation task.

\subsection{Completion and Segmentation Cooperative Module (CSCM)}
\label{sec:model_CSCM}
We utilize point proxies updated by a spatial-aware transformer decoder to generate a complete point cloud and annotate semantic segmentation labels cooperatively. It is important to highlight that point proxy features are locally feature-dominated, so we adopt the local feature up-sampling method to generate points and semantic labels.

As one point proxy contains both proxy coordination and proxy features. For point proxy features ${F} \in \mathbb{R}{^{N \times C}}$, inspired by \cite{c17}, we apply transposed convolutions as Feature Generator Module and Rebuild Head to expand feature dimension, yielding expanded features ${\bar F} \in \mathbb{R}{^{N \times S \times C'}}$, where $N$ is proxy number, $S$ is up sample factor, $C$ and $C'$ are feature dimensions. Leveraging these expanded features, a rebuilding head is used to predict point-wise offsets ${O} \in \mathbb{R}{^{N \times S \times 3}}$ from the origin query coordination ${P} \in \mathbb{R}{^{N \times 3}}$. Following this, an element-wise addition is executed to produce the final complete point coordinates ${\bar P} \in \mathbb{R}{^{NS \times 3}}$.  In parallel, a segmentation head predicts the semantic label ${L} \in \mathbb{R}{^{NS \times 1}}$ for each point.


\subsection{Training Loss}
\label{sec:model_training_loss}


To fully guide our network, we employ ground truth complete points and semantic labels as supervisory signals. For the completion task, the widely-adopted Chamfer Distance (CD) loss\cite{chamfer} is utilized to minimize the Euclidean distance between predicted points $P$ and ground truth points ${\hat P}$. For the semantic segmentation task, the variability in generated point positions means there is not a clear one-to-one mapping between $P$ and ${\hat P}$. Following \cite{c38}, for each predicted point, we identify the closest ground truth point and utilize its semantic label as ground truth. Cross-entropy loss\cite{hinton1995wake} supervises predicted logits \( L \) and ground truth labels \( {\hat L} \).

The overall loss of PointSSC consists of CD loss $\mathcal{L}_{CD}$ and cross-entropy loss $\mathcal{L}_{CE}$ with balanced parameter $\lambda$:
\begin{equation}
\mathcal{L}_{SSC}=\mathcal{L}_{CD} \left( P, {\hat P} \right) + \lambda \mathcal{L}_{ce} \left( L, {\hat L} \right).
\label{equ:loss}
\end{equation}

\begin{table*}[htbp]
  \centering
  \caption{Quantitative comparison of different methods on PointSSC dataset. CD Means Chamfer Distance(Multiplied By 1000), lower is better. F1-score@0.3 means calculating by distance threshold 0.3. The best results are in \textbf{Bold}.}
  \resizebox{\linewidth}{!}{
    \begin{tabular}{c|l|ccc|ccccccccccccccccc}
    \toprule
         & Methods & CD ($L_1$) $\downarrow$  & CD ($L_2$) $\downarrow$ & \multicolumn{1}{l|}{F1-Score@0.3 $\uparrow$} & mIoU $\uparrow$ & building & tree & road & sidewalk & person & plant & car & fence & signboard & bus & truck & streetlight & barricade & van & bicycle & motorcyclist  \\
    \midrule
    \multirow{8}[4]{*}{\begin{sideways}\rotatebox{0}{Time Splitting}\end{sideways}} & FoldingNet\cite{c14} &  610.30     &  2785.44     &  39.65\%     &   14.91    &  8.56     &  60.74     &  81.01     &  27.45     &  0.00     &  35.26     & 12.31      &  11.24     &  1.07     &  0.14     &   0.70    &  0.00     & 0.00      & 0.00      &  0.00     & 0.00 \\
          & PCN\cite{c16}   & 317.38     &  510.19     &  64.46\%     &  12.65     &  0.14    &  69.71     &  83.91     &  0.70     & 0.00      &  29.76     &  2.08     & 9.05      &  7.05     & 0.01      &  0.00     &  0.00     &  0.01     & 0.00      & 0.00      & 0.00 \\
          
          & PoinTr\cite{c18} &  3864.43     &   129947.08    &  15.66\%     &  5.21     &  0.00     &  2.69     &  0.00     &  0.00     &  0.00     &  0.00     &   2.52    &   0.00    &  0.00     & 0.00      &  0.00     &  0.00     &  0.00     & 0.00      &  0.00     &  0.00\\
          & SnowFlakeNet\cite{c17} &  461.89     &  2846.09     &  52.99\%     &  20.78     &  11.58     &  80.65     & 80.79      &  32.47     & 0.00      &  48.84     &  1.47     & 35.63      & 36.34      & 2.03      &  1.47     &  1.14     &  0.00     &  0.00     &  0.00     & 0.00 \\
          & PMP-Net++\cite{c45} &  541.82     &  3753.60     &  57.09\%     &  25.25     &  21.79     &  80.67     &  87.89     &  10.10     & 4.73      &  37.53     &  44.29     &  14.53     &  7.74     &  37.97     &  29.66     &  3.20     &  19.02     &  3.83     &  0.00     &  1.04\\
          & CasFusionNet\cite{c38} &  467.76     &  5087.18     & 70.84\%      & 45.76      & 55.20      &  90.50     &  90.11     &  55.08     &  \textbf{19.52}     & 69.91      & 59.38      &  58.46     & 45.35      &  47.89     & 37.62      & 12.53      &  9.66     &  39.85     &  13.80     & 27.37 \\
          & AdaPoinTr\cite{c19} &  237.11     &   290.71    &  78.79\%     &  45.09     &  45.53     &  87.08     &  \textbf{93.42}     & 53.36  &  13.00     &  64.46     &  55.99     &  56.41     &  42.36     &  56.44     & 37.56      &  \textbf{15.79}     &  17.61     &  43.05     &  11.31     & 28.07 \\
\cmidrule{2-22}          & \textbf{PointSSC} (Ours) & \textbf{208.94} & \textbf{248.28} & \textbf{81.42\%}  & \textbf{50.58} & \textbf{61.42} & \textbf{90.94} & 92.41 & \textbf{63.32} & 13.40  & \textbf{74.33} & \textbf{66.17} & \textbf{67.53} & \textbf{47.93} & \textbf{61.37} & \textbf{44.35} & 14.92 & \textbf{18.82} & \textbf{43.69} & \textbf{17.64} & \textbf{31.07} \\
    \midrule
    \multirow{8}[4]{*}{\begin{sideways}\rotatebox{0}{Scene Splitting}\end{sideways}} & 
    FoldingNet\cite{c14} &  1146.07     &  16950.85     & 6.06\%      &  8.11     &  0.14     &  8.79     & 84.06      &  4.08     & 0.00      & 31.60      &  1.01     &  0.00     &  0.00     & 0.00      & 0.00      &  0.00     & 0.00      &  0.00     &  0.00     & 0.00 \\
          & PCN\cite{c16}   &   809.55    &   5778.96    &  42.66\%     &    8.62   &   0.03    &    47.95   &   78.07    &   0.01    &   0.00    &    11.86   &    0.00   &   0.00    &   0.02    &   0.00    &   0.00    &   0.00    &   0.00    &    0.00   &   0.00    & 0.00 \\
          
          & PoinTr\cite{c18} & 4164.05      &   90275.69    &  15.89\%     &  0.40     &  0.00     & 0.00      &  0.02     &   0.01    &  0.00     &   0.00    &  6.06     &   0.00    & 0.00      &  0.00     &  0.24     &  0.00     &  0.00     &  0.00     &  0.00     & 0.00 \\
          & SnowFlakeNet\cite{c17} & 1055.09      &  21849.79     & 42.36\%      &  8.09     & 0.00      & 55.81      &  63.33     &  2.06     &  0.00     &  7.51     &  0.00     & 0.00      & 0.26      &  0.18    &  0.00     & 0.35      & 0.00      &   0.00    & 0.00      & 0.00 \\
          & PMP-Net++\cite{c45} & 530.80     & 3753.60      &  57.09\%     &  13.66     &  0.00     &  65.75     &  84.76     & 8.80      &   0.00    &  \textbf{16.32}     &  24.66     &  0.01     &  0.86     &  4.96     &  5.17     &  0.00     &  \textbf{0.71}     & 4.94      &  0.00     & 1.57 \\
          & CasFusionNet\cite{c38} & 664.85      &  8310.55     &  43.98\%     &    11.52   &  \textbf{0.52}     & 62.33      &  65.68     &   6.04    &  \textbf{0.08}     &  9.06     &   21.84    &  0.00     &   0.48    &   \textbf{8.08}    &  0.12     &  0.18     &  0.00     &  7.06     &  0.92     &  1.85\\
          & AdaPoinTr\cite{c19} &  493.41     &  3098.48     &  61.41\%     &  13.20     &  0.43     &  66.38     &  84.91     &  5.57     &  0.00     & 15.86      &  15.67     & 0.00      & 0.21      &  1.90     &  5.07    &  \textbf{1.55}     &  0.00     & \textbf{11.55}      &  0.06     & 2.06 \\
\cmidrule{2-22}          & \textbf{PointSSC} (Ours) &  \textbf{410.92}     & \textbf{1413.60}      &  \textbf{63.57\%}     &  \textbf{14.64}     &  0.12     &  \textbf{68.93}     &  \textbf{85.68}     &  \textbf{14.02}     &  0.00     &  13.22     &  \textbf{24.69}     &  \textbf{10.64}     & \textbf{1.78}      &  1.32     &  \textbf{5.65}     &  0.01     &  0.07     &  5.42     &  \textbf{0.54}     & \textbf{2.13} \\
    \bottomrule
    \end{tabular}%
    }
  \label{tab:experiments}%
\end{table*}%

\section{EXPERIMENTS}

\subsection{Experiment Setup}

\begin{figure}[t]
  \centering
  \includegraphics[width=0.4\textwidth]{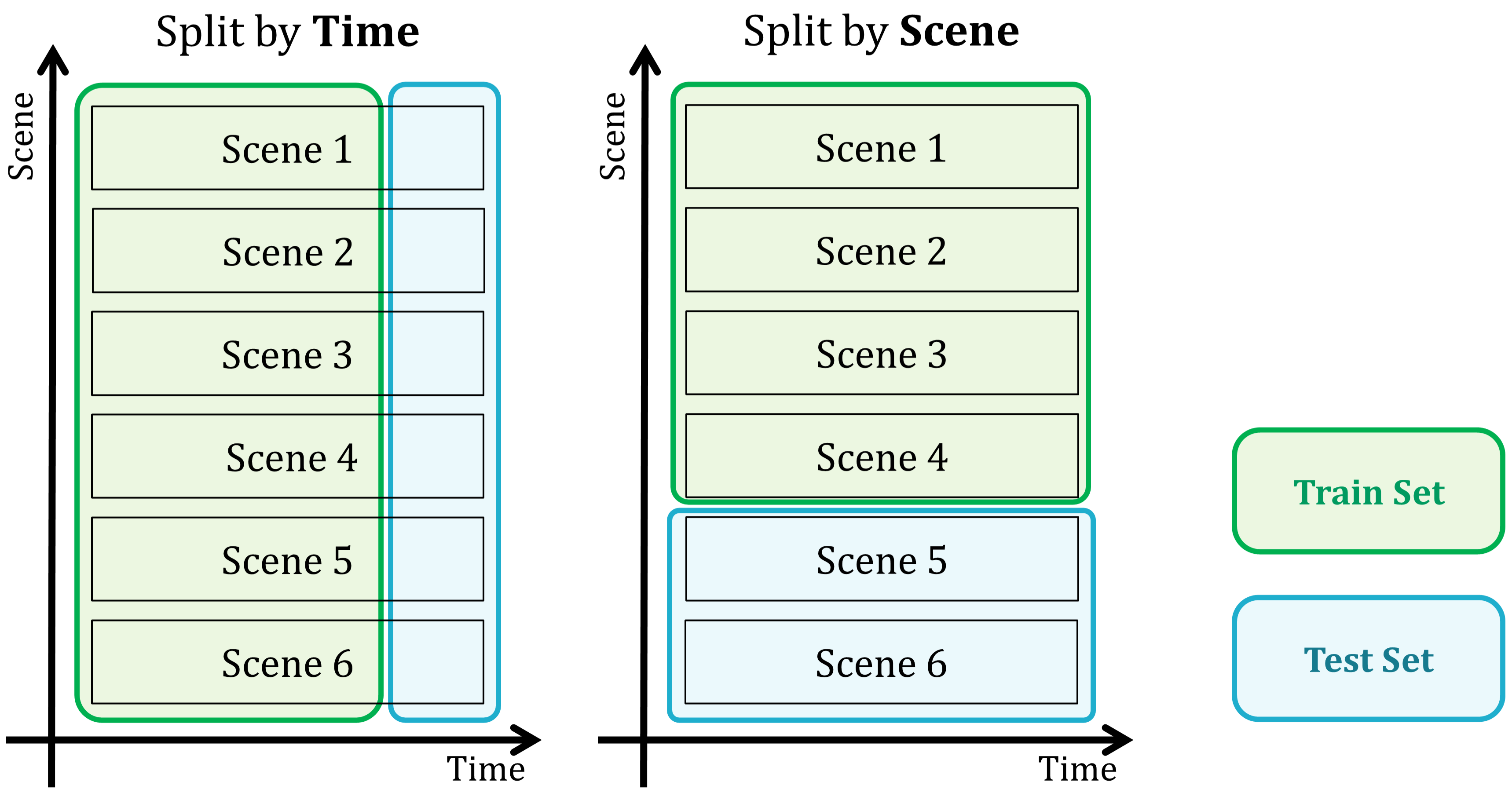}
  \caption{Two types of data division. The first one is split by time, the second one is split by scenes.}
  \label{fig:data_split}
\end{figure}

\begin{figure*}[t!]
  \centering
  \includegraphics[width=1\textwidth]{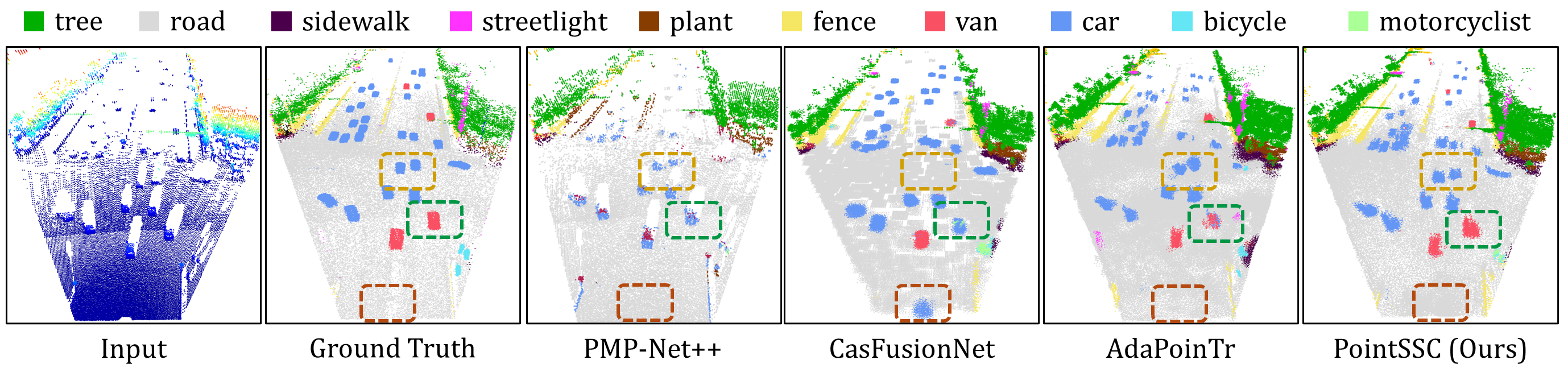}
  \caption{Visualization comparison of our model (PointSSC) with other point completion methods.}
  \label{fig:experiments}
\end{figure*}

Our PointSSC dataset is derived from six infrastructure-side intersections in V2X-Seq\cite{v2x-seq}. As in Fig. \ref{fig:data_split}, we offer two data divisions based on PointSSC. In the first division, we allocate 80\% of time-sequence frames from all six scenes for training, while the remaining 20\% is reserved for testing, which evaluates the model's capacity for expressiveness. For the second division, we employ four intersections throughout the entire time range for training and use the remaining two intersections for testing, serving to assess the model's ability to generalize across unfamiliar scenes.

\subsection{Implementation Details}

We apply Pointnet++\cite{c13} as proxy generator to extract point cloud centroids and corresponding features, which are seen as proxies. Only points within range of $\left[0m, 250m \right]$, $\left[-70m, 70m \right]$, $\left[-5m, 12m \right]$ for the X, Y, and Z axes are remained. We randomly sample 26,624 points as input and extract 832 proxies with a $\left[4, 4, 2 \right]$ downsampling ratio. The proxy generator module and CSCM use a $\left[16, 16 \right]$ upsampling factor, yielding 13,312 coarse proxies and 212,992 complete points. We set $\lambda=1$ in Equ. \ref{equ:loss} and utilize AdamW Optimizer, setting an initial learning rate of $1.0 \times 10^{-4}$ and the decay weight at $5.0 \times 10^{-4}$. All experiments run for 30 epochs with a total batch of 8 on 4 RTX A100 GPUs.

\subsection{Evaluation Metrics}
To evaluate model performance
, following \cite{c38} and \cite{c19}, we use the Chamfer Distance (CD), measured in L1-norm and L2-norm, and F1-score to evaluate the completeness of generated points. We use mean class IoU (mIoU) to evaluate the accuracy of point cloud semantic segmentation.

\subsection{Main Results}
In Tab. \ref{tab:experiments}, we show quantitative results of different networks on PointSSC. Except for \cite{c38}, other models do not have segment heads, we add CSCM mentioned in Sec. \ref{sec:model_CSCM} for a fair comparison. Our model excels in both PointSSC data division scenarios.


In Table \ref{tab:ablation_study}, we conduct ablation studies on the test set regarding the use of the spatial-aware transformer, as detailed in Sec. \ref{sec:spatial-aware_transformer}. Results indicate that solely relying on max-pooling-processed global features yields the poorest outcome. Incorporating local features enhances CD and mIoU by 5\% and 8\% respectively. The combination of both local and global features within the transformer block achieves optimal performance.

\begin{table}[t]
  \centering
  \caption{Ablation study on TEST set for spatial-aware transformer.}
    \begin{tabular}{c|l|lll}
    \toprule
          & \textbf{Ablation} & \multicolumn{1}{l}{\textbf{CD-$L_1$$\downarrow$}} & \multicolumn{1}{l}{\textbf{CD-$L_2$$\downarrow$}} & \multicolumn{1}{l}{\textbf{mIoU$\uparrow$}} \\
    \midrule
    (a)   & with global feature & 237.11 & 290.71 & 45.09 \\
    (b)   & with local feature & 225.13 & 279.92 & 49.53 \\
    (c)   & local \& global feature & \textbf{208.94} & \textbf{248.28} & \textbf{50.58} \\
    \bottomrule
    \end{tabular}%
  \label{tab:ablation_study}%
\end{table}%

\subsection{Visualization}

In Fig. \ref{fig:experiments}, we visualize the outcomes of various models on the PointSSC dataset. Our baseline outperforms all others in terms of point completion and semantic segmentation. Compared to AdaPoinTr\cite{c19}, PointSSC produces fewer noisy semantic points. Additionally, our model yields a more comprehensive and cohesive shape representation than both PMP-Net++\cite{c45} and CasFusionNet\cite{c38}.

\section{CONCLUSIONS}

In this paper, we introduce a comprehensive benchmark for point cloud semantic scene completion task, comprising a dataset and a LiDAR-based model baseline. To produce superior semantic points, we propose the spatial-aware transformer and the completion and segmentation cooperative module. Experimental results demonstrate our model's superiority over competing approaches. We earnestly hope that PointSSC will fill the blank in scene-level semantic point cloud generation datasets and draw further research interest to this domain.

\addtolength{\textheight}{0cm}   







\clearpage
\bibliographystyle{IEEEtran}
\bibliography{sample}

\end{document}